\documentclass{article}

\usepackage{microtype}
\usepackage{graphicx}
\usepackage{subfigure}
\usepackage{booktabs} \usepackage{xspace}

\usepackage{hyperref}

\everypar{\looseness=-1}
\usepackage{amssymb}

\newcommand\refsec[1]{Section~\ref{sec:#1}}
\newcommand\refsecs[2]{Sections~\ref{sec:#1} and~\ref{sec:#2}}
\newcommand\reffig[1]{Figure~\ref{fig:#1}}

\newcommand\reftab[1]{Table~\ref{tab:#1}}
\newcommand\refapp[1]{Appendix~\ref{sec:#1}}

\newcommand\refalg[1]{Algorithm~\ref{alg:#1}}

\newcommand{\ours}{\textsc{Parrot}\xspace}
 
\usepackage[accepted]{icml2020}

\usepackage[shortlabels]{enumitem}
\setlist{itemsep=0pt,parsep=2pt,topsep=3pt,partopsep=0pt}
\usepackage{amsmath}
\usepackage{nicefrac}
\usepackage{setspace}
\usepackage[skip=0pt,tableposition=top]{caption}
\usepackage[separate-uncertainty=true]{siunitx}
\usepackage{float}
\newfloat{algorithm}{t}{lop}

\expandafter\def\expandafter\normalsize\expandafter{\normalsize \setlength\abovedisplayskip{5pt}\setlength\belowdisplayskip{5pt}\setlength\abovedisplayshortskip{5pt}\setlength\belowdisplayshortskip{5pt}}

\makeatletter
\long\def\@makecaption#1#2{
\vskip 2pt
        \baselineskip 11pt
        \setbox\@tempboxa\hbox{#1. #2}
        \ifdim \wd\@tempboxa >\hsize
        \sbox{\newcaptionbox}{\small\sl #1.~}
        \newcaptionboxwid=\wd\newcaptionbox
        \usebox\newcaptionbox {\footnotesize #2}
\else 
          \centerline{{\small\sl #1.} {\small #2}} 
        \fi}

\def\section{\@startsection{section}{1}{\z@}{-0.06in}{0.01in}
             {\large\bf\raggedright}}
\def\subsection{\@startsection{subsection}{2}{\z@}{-0.06in}{0.01in}
                {\normalsize\bf\raggedright}}
\def\paragraph{\@startsection{paragraph}{4}{\z@}{0.5ex plus
  0.2ex minus .2ex}{-1em}{\normalsize\bf}}
\makeatother

\icmltitlerunning{An Imitation Learning Approach for Cache Replacement}

\begin{document}

\twocolumn[
\icmltitle{An Imitation Learning Approach for Cache Replacement}

\begin{icmlauthorlist}
\icmlauthor{Evan Zheran Liu}{stanford,google}
\icmlauthor{Milad Hashemi}{google}
\icmlauthor{Kevin Swersky}{google}
\icmlauthor{Parthasarathy Ranganathan}{google}
\icmlauthor{Junwhan Ahn}{google}
\end{icmlauthorlist}

\icmlaffiliation{stanford}{Department of Computer Science, Stanford University, California, USA}
\icmlaffiliation{google}{Google Research, Sunnyvale, California, USA}

\icmlcorrespondingauthor{Evan Z. Liu}{evanliu@cs.stanford.edu}

\icmlkeywords{Machine Learning, ICML}

\vskip 0.3in
]

\printAffiliationsAndNotice{}  

\begin{abstract}
    Program execution speed critically depends on increasing cache hits,
    as cache hits are orders of magnitude faster than misses.
    To increase cache hits, we focus on the problem of cache replacement:
    choosing which cache line to evict upon inserting a new line.
    This is challenging because it requires planning far ahead and currently there is no known practical solution.
    As a result,
    current replacement policies typically resort to heuristics designed for specific common access patterns,
    which fail on more diverse and complex access patterns.
    In contrast, we propose an imitation learning approach to automatically learn cache access patterns by leveraging Belady's,
    an oracle policy that computes the optimal eviction decision given the future cache accesses.
    While directly applying Belady's is infeasible since the future is unknown,
    we train a policy conditioned only on \emph{past} accesses that accurately approximates Belady's even on diverse and complex access patterns, and call this approach \ours.
    When evaluated on 13 of the most memory-intensive SPEC applications,
    \ours increases cache miss rates by 20\% over the current state of the art.
    In addition, on a large-scale web search benchmark, \ours increases cache hit rates by 61\% over a conventional LRU policy.
    We release a Gym environment to facilitate research in this area, as data is plentiful,
    and further advancements can have significant real-world impact.
\end{abstract}
 \section{Introduction}\label{sec:intro}

Caching is a universal concept in computer systems that bridges the performance gap between different levels of data storage hierarchies, found everywhere from databases to operating systems to CPUs~\citep{jouppi1990improving, harty1992application, xu2013characterizing, cidon2016cliffhanger}.
Correctly selecting what data is stored in caches is critical for latency,
as accessing the data directly from the cache (a \emph{cache hit}) is orders of magnitude faster than retrieving the data from a lower level in the storage hierarchy (a \emph{cache miss}).
For example, \citet{cidon2016cliffhanger} show that improving cache hit rates of web-scale applications by just 1\% can decrease total latency by as much as 35\%.

Thus, general techniques for increasing cache hit rates would significantly improve performance at all levels of the software stack.
Broadly, two main avenues for increasing cache hit rates exist:
(i)~avoiding future cache misses by proactively prefetching the appropriate data into the cache beforehand;
and (ii)~strategically selecting which data to evict from the cache when making space for new data (cache replacement).
Simply increasing cache sizes is a tempting third avenue, but is generally prohibitively expensive.

\begin{figure}\center
    \includegraphics[width=\linewidth]{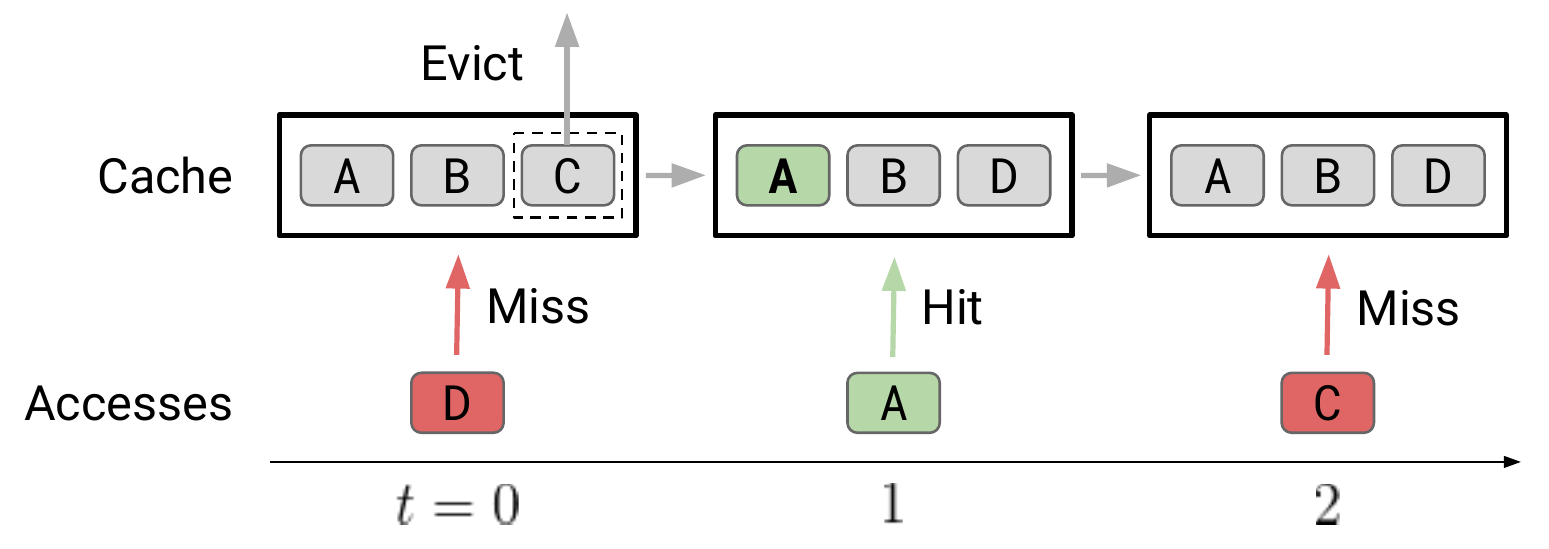}
    \caption{
        Cache replacement.
At $t = 0$, line D is accessed, causing a cache miss.
        The replacement policy chooses between lines A, B, and C in the cache and in this case evicts C.
        At $t = 1$, line A is accessed and is already in the cache, causing a cache hit.
        No action from the replacement policy is needed.
        At $t = 2$, line C is accessed, causing another cache miss.
        The replacement policy could have avoided this miss by evicting a different line at $t = 0$.
}
    \label{fig:problem_setup}
\end{figure}

This work focuses on single-level cache replacement (\reffig{problem_setup}).
When a new block of data (referred to as a \emph{line}) is added to the cache (i.e., due to a cache miss), an existing cache line must be evicted from the cache to make space for the new line.
To do this, during cache misses, a cache replacement policy takes as inputs the currently accessed line and the lines in the cache and outputs which of the cache lines to evict.

Prior work frequently relies on manually-engineered heuristics to capture the most common cache access patterns,
such as evicting the most recently used (MRU) or least recently used (LRU) cache lines,
or trying to identify the cache lines that are frequently reused vs.\ those that are not~\cite{qureshi2007adaptive, jaleel2010high, jain2016back, shi2019applying}.
These heuristics perform well on the specific simple access patterns they target,
but they only target a small fraction of all possible access patterns,
and consequently they perform poorly on programs with more diverse and complex access patterns.
Current cache replacement policies resort to heuristics as practical theoretical foundations have not yet been developed~\citep{beckmann2017maximizing}.

We propose a new approach for learning cache replacement policies
by leveraging Belady's optimal policy~\citep{belady1966study} in the framework of imitation learning (IL), and name this approach \ours.\footnote{Parrots are known for their ability to \emph{imitate} others.} Belady's optimal policy (Belady's for short) is an oracle policy that computes the theoretically
optimal cache eviction decision based on knowledge of future cache accesses,
which we propose to approximate with a policy that only conditions on the \emph{past} accesses.
While our main goal is to establish (imitation) learned replacement policies as a proof-of-concept,
we note that deploying such learned policies requires solving practical challenges, e.g., model latency may overshadow gains due to better cache replacement.
We address some of these challenges in \refsec{practical} and highlight promising future directions in \refsec{discussion}.

Hawkeye~\citep{jain2016back} and Glider~\citep{shi2019applying} were the first to propose learning from Belady's. 
They train a binary classifier to predict if a cache line will soon be reused (cache-friendly) or not (cache-averse),
evicting the cache-averse lines before the cache-friendly ones and relying on a traditional heuristic to determine 
which lines are evicted first within the cache-friendly and cache-averse groups.
Training such a binary classifier avoids the challenges (e.g., \emph{compounding errors}) of directly learning a policy,
but relying on the traditional heuristic heavily limits the expressivity of the policy class that these methods optimize over,
which prevents them from accurately approximating Belady's.
In contrast,
our work is the first to propose cache replacement as an IL problem,
which allows us to directly train a replacement policy end-to-end over a much more expressive policy class to approximate Belady's.
This represents a novel way of leveraging Belady's and provides a new framework for learning end-to-end replacement policies.

Concretely, this paper makes the following contributions:

\begin{itemize}
    \item We cast cache replacement as an imitation learning problem, leveraging Belady's in a new way (\refsec{mdp}).
    \item We develop a neural architecture for end-to-end cache replacement and several supervised tasks that further improve its performance over standard IL (\refsec{approach}).
    \item Our proposed approach, \ours, exceeds the state-of-the-art replacement policy's hit rates by over 20\% on memory-intensive CPU benchmarks.
    On an industrial-scale web search workload, \ours improves cache hit rates by 61\% over a commonly implemented LRU policy (\refsec{experiments}).
    \item We propose cache replacement as a challenging new IL\slash RL (reinforcement learning) benchmark involving dynamically changing action spaces, delayed rewards, and significant real-world impact. To that end, we release an associated Gym environment (\refsec{discussion}).
\end{itemize}

 \section{Cache Preliminaries}\label{sec:preliminaries}
We begin with cache preliminaries before formulating cache replacement as learning a policy over a Markov decision process in \refsec{mdp}.
We describe the details relevant to CPU caches, which we evaluate our approach on, but as caching is a general concept, our approach can be extended towards other cache structures as well.

A cache is a memory structure that maintains a portion of the data from a larger memory. If the desired data is located in the cache when it is required, this is advantageous, as smaller memories are faster to access than larger memories. Provided a memory structure, there is a question of how to best organize it into a cache. In CPUs, caches operate in terms of atomic blocks of memory or \emph{cache lines} (typically 64-bytes large). This is the minimum granularity of data that can be accessed from the cache.

During a memory access, the cache must be searched for the requested data.
\emph{Fully-associative} caches layout all data in a single flat structure,
but this is generally prohibitively expensive, as locating the requested data requires searching through all data in the cache.
Instead, CPU caches are often $W$-way \emph{set-associative} caches of size $N \times W$, consisting of $N$ \emph{cache sets}, where each cache set holds $W$ \emph{cache lines} $\{l_1, \ldots, l_W\}$.
Each line maps to a particular cache set (typically determined by the lower order bits of line's address),
so only the $W$ lines within that set must be searched.

During execution, programs read from and write to \emph{memory addresses} by executing load or store instructions. These load\slash store instructions have unique identifiers known as \emph{program counters} (PCs).
If the address is located in the cache, this is called a \emph{cache hit}.
Otherwise, this is a \emph{cache miss}, and the data at that address must be retrieved from a larger memory.
Once the data is retrieved, it is generally added to the appropriate cache set (as recently accessed lines could be accessed again).
Since each cache set can only hold $W$ lines, if a new line is added to a cache set already containing $W$ lines, the cache replacement policy must choose an existing line to replace.
This is called a \emph{cache eviction} and selecting the optimal line to evict is the cache replacement problem.

\paragraph{Belady's Optimal Policy.}
Given knowledge of future cache accesses, Belady's computes the \emph{optimal} cache eviction decision.
Specifically, at each timestep $t$, Belady's computes the \emph{reuse distance} $d_t(l_w)$ for each line $l_w$ in the cache set,
which is defined as the number of total cache accesses until the next access to $l_w$.
Then, Belady's chooses to evict the line with the highest reuse distance, effectively the line used furthest in the future,
i.e., $\arg\max_{w = 1, \ldots, W}d_t(l_w)$.

\section{Casting Cache Replacement as Imitation Learning}\label{sec:mdp}

We cast cache replacement as learning a policy on an episodic Markov decision process $\langle \mathcal{S}, \mathcal{A}_s, R, P\rangle$ in order to leverage techniques from imitation learning.
Specifically, the state at the $t$-th timestep $s_t = (s^c_t, s^a_t, s^h_t) \in \mathcal{S}$ consists of three components, where:
\begin{itemize}
    \item $s^a_t = (m_t, pc_t)$ is the current cache access, consisting of the currently accessed cache line address $m_t$ and the unique program counter $pc_t$ of the access.
    
    \item $s^c_t = \{l_1, \dots, l_W\}$ is the cache state consisting of the $W$ cache line addresses currently in the cache set accessed by $s^a_t$ (the replacement policy does not require the whole cache state including other cache sets to make a decision).\footnote{A cache set can have less than $W$ cache lines for the first $W-1$ cache accesses (small fraction of program execution). In this case, no eviction is needed to insert the line.}
    
    \item $s^h_t = (\{m_1, \dots, m_{t-1}\}, \{pc_1, \dots, pc_{t-1}\})$ is the history of all past cache accesses.
    In practice, we effectively only condition on the past $H$ accesses.
\end{itemize}

The action set $\mathcal{A}_{s_t}$ available at a state $s_t = (s^c_t, s^a_t, s^h_t)$ is defined as follows:
During cache misses, i.e., $m_t \not\in s^c_t$, the action set $\mathcal{A}_{s_t}$ consists of the integers $\{1, \dots, W\}$, where action $w$ corresponds to evicting line $l_w$.
Otherwise, during cache hits, the action set $\mathcal{A}_{s_t}$ consists of a single no-op action $a_{\textrm{no-op}}$, since no line must be evicted.

The transition dynamics $P(s_{t + 1} \mid a_t, s_t)$ are given by the dynamics of the three parts of the state.
The dynamics of the next cache access $s^a_{t + 1}$ and the cache access history $s^h_{t + 1}$ are independent of the action $a_t$ and are defined by the program being executed.
Specifically, the next access $s^a_{t + 1} = (m_{t + 1}, pc_{t + 1})$ is simply the next memory address the program accesses and its associated PC.
The $t$-th access is appended to $s^h_{t + 1}$, i.e., $s^h_{t + 1} = (\{m_1, \dots, m_{t-1}, m_t\}, \{pc_1, \dots, pc_{t-1}, pc_t\})$.

The dynamics of the cache state are determined by the actions taken by the replacement policy.
At state $s_t$ with $s^c_t = \{l_1, \dots, l_W\}$ and $s^a_t = (m_t, pc_t)$:
A cache hit does not change the cache state, i.e., $s^c_{t + 1} = s^c_t$, as the accessed line is already available in the cache.
A cache miss replaces the selected line with the newly accessed line, i.e., 
$s^c_{t + 1} = \{l_1, \dots, l_{w - 1}, l_{w + 1}, \dots, l_W, m_t\}$ where $a_t = w$.

The reward $R(s_t)$ is $0$ for a cache miss (i.e., $m_t \not\in s^c_t$) and is 1 otherwise for a cache hit.
The goal is to learn a policy $\pi_\theta(a_t \mid s_t)$ that maximizes the undiscounted total number of cache hits (the reward), $\sum_{t = 0}^T R(s_t)$, for a sequence of $T$ cache accesses $(m_1, pc_1), \dots, (m_T, pc_T)$.

In this paper, we formulate this task as an imitation learning problem.
During training,
we can compute the optimal policy (Belady's) $\pi^*(a_t \mid s_t, (m_{t + 1}, pc_{t + 1}), \dots, (m_T, pc_T))$,
by leveraging that the future accesses are fixed.
Then, our approach learns a policy $\pi_\theta(a_t \mid s_t)$ to approximate the optimal policy without using the future accesses, as future accesses are unknown during test time.

\begin{figure}\center
    \includegraphics[width=0.96\linewidth,trim={0 0 0 1cm},clip]{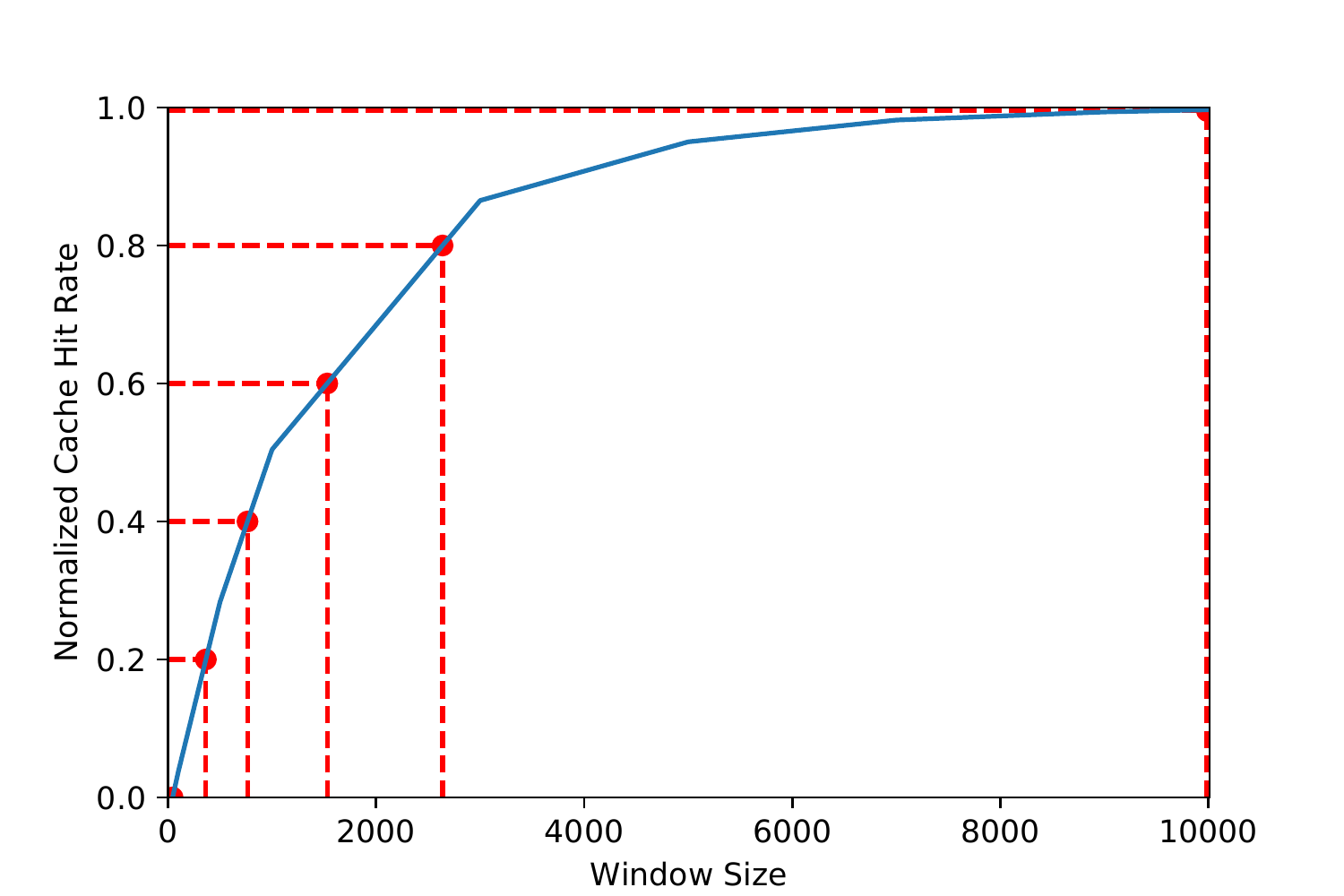}
    \caption{
        Normalized cache hit rates of Belady's vs.\ the number of accesses it looks into the future.
        Achieving 80\% the performance of Belady's with an infinite window size requires accurately computing reuse distances for lines 2600 accesses into the future. 
    }
    \label{fig:belady_window_size}
\end{figure}

To demonstrate the difficulty of the problem, \reffig{belady_window_size} shows the amount of future information required to match the performance of Belady's on a common computer architecture benchmark (\textit{omnetpp}, Section \ref{sec:experiments}).
We compute this by imposing a future window of size $x$ on Belady's, which we call $\textrm{Belady}_x$,
Within the window ($\text{reuse distances} \leq x$), $\textrm{Belady}_x$ observes exact reuse distances,
and sets the reuse distances of the remaining cache lines (with $\text{reuse distance} > x$) to $\infty$.
Then, $\textrm{Belady}_x$ evicts the line with the highest reuse distance, breaking ties randomly.
The cache hit rate of $\textrm{Belady}_x$ is plotted on the y-axis, normalized so that 0 and 1 correspond to the cache hit rate of LRU and $\textrm{Belady}_\infty$ (the normal unconstrained version of Belady's), respectively.
As the figure shows, a significant amount of future information is required to fully match Belady's performance.

 \section{\ours: Learning to Imitate Belady's}\label{sec:approach}

\subsection{Model and Training Overview}

\begin{figure}\center
    \includegraphics[width=0.98\linewidth]{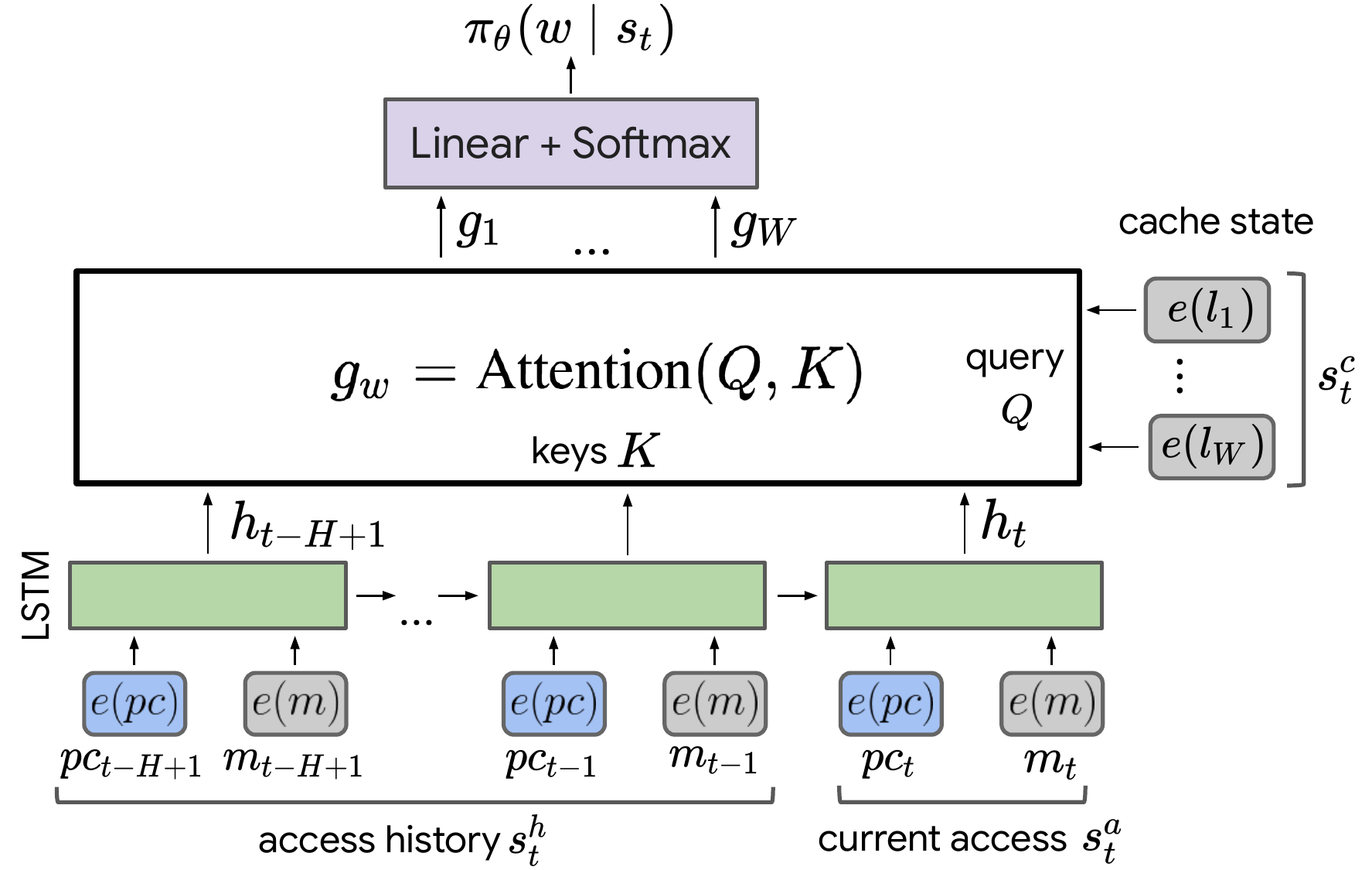}
    \caption{
        Neural architecture of \ours.
    }
    \label{fig:model}
\end{figure}

\paragraph{Model.} Below, we overview the basic architecture of the \ours policy $\pi_\theta(a_t \mid s_t)$ (\reffig{model}),
which draws on the Transformer~\citep{vaswani2017attention} and BiDAF~\cite{seo2016bidirectional} architectures.
See \refapp{architecture_details} for the full details.

\begin{enumerate}
    \item Embed the current cache access $s^a_t = (m_t, pc_t)$ to obtain memory address embedding $e(m_t)$ and PC embedding $e(pc_t)$ and pass them through an LSTM to obtain cell state $c_t$ and hidden state $h_t$:
        \begin{equation*}
            c_t, h_t = \mathrm{LSTM}([e(m_t); e(pc_t)], c_{t - 1}, h_{t - 1})
        \end{equation*}
        
    \item Keep the past $H$ hidden states, $[h_{t - H + 1}, \ldots, h_t]$, representing an embedding of the cache access history $s^h_t$ and current cache access $s^a_t$.
    
    \item Form a context $g_w$ for each cache line $l_w$ in the cache state $s^c_t$ by embedding each line as $e(l_w)$ and attending over the past $H$ hidden states with $e(l_w)$:
        \begin{gather*}
            g_w = \textrm{Attention}(Q, K) \\
            \text{where } \text{query }Q = e(l_w), \text{keys }K = [h_{t - H + 1}, \ldots, h_t]
        \end{gather*}

    \item Apply a final dense layer and softmax on top of these line contexts to obtain the policy:
        \begin{equation*}
            \pi_\theta(a_t = w \mid s_t) = \textrm{softmax}(\textrm{dense}(g_w))
        \end{equation*}
        
    \item Choose $\arg\max_{a\in \mathcal{A}_{s_t}}{\pi_\theta(a \mid s_t)}$ as the replacement action to take at timestep $t$.
\end{enumerate}

\paragraph{Training.} \refalg{training} summarizes the training algorithm for the \ours policy $\pi_\theta$. The high-level strategy is to visit a set of states $B$ and then update the parameters $\theta$ to make the same eviction decision as the optimal policy $\pi^*$ on each state $s \in B$ via the loss function $\mathcal{L}_\theta(s, \pi^*)$.

\begin{algorithm}
    \small
    \setstretch{1.1}
    \begin{flushleft}
    \begin{algorithmic}[1]
        \STATE Initialize policy $\pi_\theta$
        \FOR{$\textrm{step} = 0$ \textbf{to} $K$}
            \IF{$\textrm{step} \equiv 0 \pmod{5000}$}
                \STATE Collect data set of visited states $B = \{s_t\}_{t = 0}^T$ by following $\pi_\theta$ on all accesses $(m_1, pc_1), \dots, (m_T, pc_T)$
            \ENDIF
            
            \STATE Sample contiguous accesses $\{s_{t}\}_{t = l - H}^{l + H}$ from $B$
            \STATE Warm up policy $\pi_\theta$ on initial $H$ accesses $(m_{l - H}, pc_{l - H}), \dots, (m_l, pc_l)$
            \STATE Compute loss $\mathcal{L} = \sum_{t = l}^{l + H} \mathcal{L}_\theta(s_t, \pi^*)$
            \STATE Update policy parameters $\theta$ based on loss $\mathcal{L}$
        \ENDFOR
    \end{algorithmic}
    \end{flushleft}
    \caption{\ours training algorithm}
    \label{alg:training}
\end{algorithm}

First, we convert a given sequence of consecutive cache accesses $(m_1, pc_1), \dots, (m_T, pc_T)$ into states $s_0, \dots, s_T$ (\refsec{compounding_errors}), on which we can compute the optimal action with Belady's (lines 3--5).
Given the states, we train \ours with truncated backpropagation through time (lines 6--9).
We sample batches of consecutive states $s_{l - H}, \dots, s_{l + H}$ and
initialize the LSTM hidden state of our policy on the cache accesses of $s_{l - H}$ to $s_{l - 1}$.
Then, we apply our replacement policy $\pi_\theta$ to each of the remaining states $s_l, \dots, s_{l + H - 1}$ in order to compute the loss $\mathcal{L}_\theta(s_t, \pi^*)$ (\refsecs{ranking_loss}{reuse_distance}),
which encourages the learned replacement policy to make the same decisions as Belady's.

\subsection{Avoiding Compounding Errors}\label{sec:compounding_errors}

Since we are only given the cache accesses and not the states,
we must determine which replacement policy to follow on these cache accesses to obtain the states $B$.
Naively, one natural policy to follow is the optimal policy $\pi^*$.
However, this leads to \emph{compounding errors}~\citep{ross2011reduction, daume2009search, bengio2015scheduled},
where the distribution of states seen during test time (when following the learned policy) differs from the distribution of states seen during training
(when following the oracle policy).
At test time, since \ours learns an imperfect approximation of the oracle policy,
it will eventually make a mistake and evict a suboptimal cache line.
This leads to cache states that are different from those seen during training, which the learned policy has not trained on, leading to further mistakes.

To address this problem, we leverage the DAgger algorithm~\citep{ross2011reduction}.
DAgger avoids compounding errors by also following the current learned policy $\pi_\theta$
instead of the oracle policy $\pi^*$ to collect $B$ during training,
which forces the distribution of training states to match that of test states.
As \ours updates the policy,
the current policy becomes increasingly different from the policy used to collect $B$,
causing the training state distribution $B$ to drift from the test state distribution.
To mitigate this, we periodically update $B$ every 5000 parameter updates by recollecting $B$ again under the current policy.
Based on the recommendation in \citep{ross2011reduction}, we follow the oracle policy the first time we collect $B$, since at that point, the policy $\pi_\theta$ is still random and likely to make poor eviction decisions.

Notably, this approach is possible because we can compute our oracle policy (Belady's) at any state during training, as long as the future accesses are known.
This differs from many IL tasks~\citep{hosu2016playing, vecerik2017leveraging}, where querying the expert is expensive and limited.

\subsection{Ranking Loss}\label{sec:ranking_loss}

Once the states $B$ are collected,
we update our policy $\pi_\theta$ to better approximate Belady's $\pi^*$ on these states via the loss function $\mathcal{L}_\theta(s, \pi^*)$.
A simple log-likelihood (LL) behavior cloning loss~\citep{pomerleau1989alvinn} $\mathcal{L}_\theta(s, \pi^*) = \log \pi_\theta(\pi^*(s) \mid s)$ encourages the learned policy to place probability mass on the optimal action $\pi^*(s)$.
However, in the setting where the \emph{distribution} $\pi^*(a \mid s)$ is known, instead of just the optimal action $\pi^*(s)$,
optimizing to match this distribution can provide more supervision,
similar to the intuition of distillation~\citep{hinton2015distilling}.
Thus, we propose an alternate ranking loss to leverage this additional supervision.

Concretely, \ours uses a differentiable approximation~\citep{qin2010general} of normalized discounted cumulative gain (NDCG) with reuse distance as the relevancy metric:
\begin{align*}
    \mathcal{L}^{\textrm{rank}}_\theta(s_t, \pi^*) &= -\frac{\mathrm{DCG}}{\mathrm{IDCG}} \\
    \textrm{where } \mathrm{DCG} &= \sum_{w = 1}^W \frac{d_t(l_w) - 1}{\log(\textrm{pos}(l_w) + 1)} \\
    \textrm{pos}(l_w) &= \sum_{i \neq w} \sigma(-\alpha(\pi_\theta(i \mid s_t) - \pi_\theta(w \mid s_t)).
\end{align*}
Here, $\textrm{pos}(l_w)$ is a differentiable approximation of the rank of line $l_w$,
ranked by how much probability the policy $\pi_\theta$ places on evicting $l_w$,
where $\alpha = 10$ is a hyperparameter and $\sigma$ is the sigmoid function.
$\mathrm{IDCG}$ is a normalization constant set so that $-1 \leq \mathcal{L}_\theta^{\textrm{rank}} \leq 0$,
equal to the value of $\mathrm{DCG}$ when the policy $\pi_\theta$ correctly places probability mass on the lines in descending order of reuse distance.
This loss function improves cache hit rates by heavily penalizing $\pi_\theta$ for placing probability on lines with low reuse distance, which will likely lead to cache misses,
and only lightly penalizing $\pi_\theta$ for placing probability on lines with higher reuse distance,
which are closer to being optimal and are less likely to lead to cache misses.

Optimizing our loss function is similar to optimizing the Kullback-Liebler (KL) divergence~\citep{kullback1951information} between a smoothed version of Belady's,
which evicts line $l_w$ with probability proportional to its exponentiated reuse distance $e^{d_t(l_w)}$, and our policy $\pi_\theta$.
Directly optimizing the KL between the non-smoothed oracle policy and our policy just recovers the normal LL loss, since Belady's actually places all of its probability on a single line.

\subsection{Predicting Reuse Distance}\label{sec:reuse_distance}

To add further supervision during training,
we propose to predict the reuse distances of each cache line as an auxiliary task~\citep{jaderberg2016reinforcement,mirowski2016learning,lample2017playing}.
Concretely, we add a second fully-connected head on \ours's network that takes as inputs the per-line context embeddings $g_w$ and outputs predictions of the log-reuse distance $\hat{d}(g_w)$.
We train this head with a mean-squared error loss $\mathcal{L}^{\textrm{reuse}}_\theta(s, \pi^*) = \frac{1}{W}\sum_{w = 1}^W (\hat{d}(g_w) - \log d_t(l_w))^2$.
Intuitively,
since the reuse distance predicting head shares the same body as the policy head $\pi_\theta$,
learning to predict reuse distances helps learn better representations in the rest of the network.
Overall, we train our policy with loss $\mathcal{L}_\theta(s, \pi^*) = \mathcal{L}^{\textrm{rank}}_\theta(s, \pi^*) + \mathcal{L}^{\textrm{reuse}}_\theta(s, \pi^*)$.

\subsection{Towards Practicality}\label{sec:practical}

\begin{figure}\center
    \includegraphics[width=0.8\linewidth]{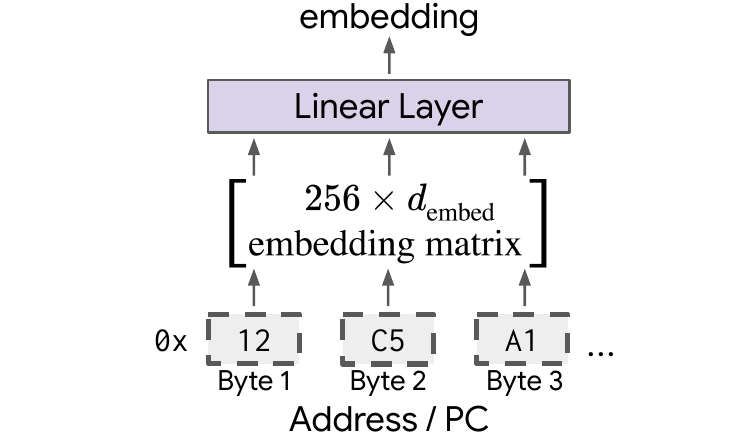}
    \caption{
        Byte embedder, taking only a few kilobytes of memory.
    }
    \label{fig:byte_embedder}
\end{figure}

The goal of this work is to establish directly imitating Belady's as a proof-of-concept.
Applying approaches like \ours to real-world systems requires reducing model size and latency to prevent overshadowing improved cache replacement.
We leave these challenges to future work, but highlight one way to reduce model size in this section, and discuss further promising directions in \refsec{discussion}.

In the full-sized \ours model, we learn a separate embedding for each PC and memory address, akin to word vectors~\citep{mikolov2013distributed} in natural language processing.
While this approach performs well, these embeddings can require tens of megabytes to store for real-world programs that access hundreds of thousands of unique memory addresses.

To reduce model size, we propose learning a byte embedder shared across all memory addresses, only requiring several kilobytes of storage.
This byte embedder embeds each memory address (or PC) by embedding each byte separately and then passing a small linear layer over their concatenated outputs (\reffig{byte_embedder}).
In principle, this can learn a hierarchical representation, that separately represents large memory regions (upper bytes of an address) and finer-grained objects (lower bytes). \section{Experiments}\label{sec:experiments}

\subsection{Experimental Setup}

Following \citet{shi2019applying}, we evaluate our approach on a three-level cache hierarchy with a 4-way \SI{32}{KB} L1 cache, a 8-way \SI{256}{KB} L2 cache, and a 16-way \SI{2}{MB} last-level cache. We apply our approach to the last-level cache while using the LRU replacement policy for L1/L2 caches.

For benchmark workloads, we evaluate on the memory-intensive SPEC CPU2006~\citep{spec} applications used by \citet{shi2019applying}.
In addition, we evaluate on Google Web Search, an industrial-scale application that serves billions of queries per day, to further evaluate the effectiveness of \ours on real-world applications with complex access patterns and large working sets.

For each of these programs, we run them and collect raw memory access traces over a 50 second interval using dynamic binary instrumentation tools~\cite{ bruening2003infrastructure}. This produces the sequence of all memory accesses that the program makes during that interval. Last-level cache access traces are obtained from this sequence by passing the raw memory accesses through the L1 and L2 caches using an LRU replacement policy.

As this produces a large amount of data, we then sample the resultant trace for our training data~\citep{qureshi2007adaptive}. We randomly choose 64 sets and collect the accesses to those sets on the last-level cache,
totaling an average of about 5M accesses per program.
Concretely, this yields a sequence of accesses $(m_1, pc_1), ..., (m_T, pc_T)$.
We train replacement policies on the first 80\% of this sequence, validate on the next 10\%, and report test results on the final 10\%.

Our evaluation focuses on two key metrics representing the efficiency of cache replacement policies.
First, as increasing cache hit rates is highly correlated to decreasing program latency~\cite{qureshi2007adaptive, shi2019applying, jain2016back}, we evaluate our policies using raw cache hit rates.
Second, we report \emph{normalized cache hit rates}, representing the gap between LRU (the most common replacement policy) and Belady's (the optimal replacement policy).
For a policy with hit rate $r$, we define the normalized cache hit rate as
$\frac{(r - r_\mathrm{LRU})}{(r_\mathrm{opt} - r_\mathrm{LRU})}$, where $r_\mathrm{LRU}$ and $r_\mathrm{opt}$ are the hit rates of LRU and Belady's, respectively.
The normalized hit rate represents the effectiveness of a given policy with respect to the two baselines, LRU (normalized hit rate of 0) and Belady's (normalized hit rate of 1).

We compare the following four approaches:
\begin{enumerate}
    \item \ours: trained with the full-sized model, learning a separate embedding for each PC and address.
    \item \ours (byte): trained with the much smaller byte embedder (\refsec{practical}).
    \item Glider~\citep{shi2019applying}: the state-of-the-art cache replacement policy, based on the results reported in their paper.
    \item Nearest Neighbor: a nearest neighbors version of Belady's, which finds the longest matching PC and memory address suffix in the training data and follows the Belady's decision of that.
\end{enumerate}

The SPEC2006 program accesses we evaluate on may slightly differ from those used by \citet{shi2019applying} in evaluating Glider, as the latter is not publicly available.
However, to ensure a fair comparison, we verified that the measured hit rates for LRU and Belady's on our cache accesses are close to the numbers reported by \citet{shi2019applying}, and we only compare on normalized cache hit rates.
Since Glider's hit rates are not available on Web Search, we compare \ours against LRU, the policy frequently used in production CPU caches.
The reported hit rates for \ours, LRU, Belady's, and Nearest Neighbors are measured on the test sets.
We apply early stopping on \ours, based on the cache hit rate on the validation set.
For \ours, we report results averaged over 3 random seeds,
using the same minimally-tuned hyperparameters in all domains.
These hyperparameters were
tuned exclusively on the validation set of omnetpp (full details in \refapp{experiment_details}).

\subsection{Main Results}\label{sec:main_results}

\begin{table*}
    \center\small
    \caption{
        Raw cache hit rates. Optimal is the hit rate of Belady's.
        Averaged over all programs, \ours (3 seeds) outperforms LRU by 16\%.
    }
    \label{tab:raw_cache_hit_rates}
    \resizebox{\textwidth}{!}{\renewcommand{\arraystretch}{1.2}
    \begin{tabular}{ccccccccccccccc}
        \toprule
         & \textbf{astar} & \textbf{bwaves} & \textbf{bzip} & \textbf{cactusadm} &
         \textbf{gems} & \textbf{lbm} & \textbf{leslie3d} & \textbf{libq}
         &\textbf{mcf} & \textbf{milc} & \textbf{omnetpp} & \textbf{sphinx3} &
         \textbf{xalanc} & \textbf{Web Search} \\
        \cmidrule(r){2-14} \cmidrule(l){15-15}
        Optimal & 43.5\% & 8.7\% & 78.4\% & 38.8\% & 26.5\% & 31.3\% & 31.9\% & 5.8\% & 46.8\% & 2.4\% & 45.1\% & 38.2\% & 33.3\% & 67.5\% \\
        LRU & 20.0\% & 4.5\% & 56.1\% & 7.4\% & 9.9\% & 0.0\% & 12.7\% & 0.0\% & 25.3\% & 0.1\% & 26.1\% & 9.5\% & 6.6\% & 45.5\% \\
        \ours & 34.4\% & 7.8\% & 64.5\% & 38.6\% & 26.0\% & 30.8\% & 31.7\% & 5.4\% & 41.4\% & 2.1\% & 41.4\% & 36.7\% & 30.4\% & 59.0\% \\
        \bottomrule
    \end{tabular}
    }
\end{table*}

\begin{figure*}\center
    \includegraphics[width=\linewidth]{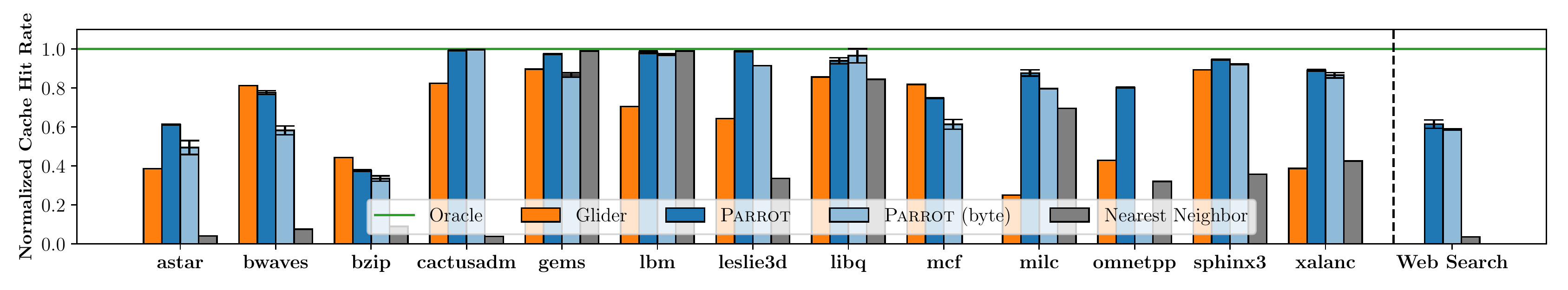}
    \caption{
        Comparison of \ours with the state-of-the-art replacement policy, Glider.
        We evaluate two versions of \ours, the full-sized model (\ours) and the byte embedder model (\ours (byte)), and report the mean performance over 3 seeds with 1-standard deviation error bars.
        On the SPEC2006 programs (left), \ours with the full-sized model improves hit rates over Glider by 20\% on average.
    }
    \label{fig:main_results}
\end{figure*}

\reftab{raw_cache_hit_rates} compares the raw cache hit rate of \ours with that of Belady's and LRU.
\ours achieves significantly higher cache hit rates than LRU on every program, ranging from 2\% to 30\%.
Averaged over all programs, \ours achieves 16\% higher cache hit rates than LRU.
According to prior study on cache sensitivity of SPEC2006 workloads~\citep{jaleel2010memory}, achieving the same level of cache hit rates as \ours with LRU would require increasing the cache capacity by 2--3x (e.g., omnetpp and mcf) to 16x (e.g., libquantum).

On the Web Search benchmark, \ours achieves a 61\% higher normalized cache hit rate and 13.5\% higher raw cache hit rate than LRU, demonstrating \ours's practical ability to scale to the complex memory access patterns found in datacenter-scale workloads.

\reffig{main_results} compares the normalized cache hit rates of \ours and Glider.
With the full-sized model, \ours outperforms Glider on 10 of the 13 SPEC2006 programs, achieving a 20\% higher normalized cache hit rate averaged over all programs;
on the remaining 3 programs (bzip, bwaves, and mcf), Glider performs marginally better.
Additionally, \ours achieves consistent performance with low variance across seeds.

\paragraph{Reducing model size.}
Though learning \ours from scratch with the byte embedder does not perform as well as the full-sized model, the byte embedder model is significantly smaller and still achieves an average of 8\% higher normalized cache hit rate than Glider (\reffig{main_results}).
In \refsec{discussion}, we highlight promising future directions to reduce the performance gap and further reduce model size and latency.

\paragraph{Generalization.}
An effective cache replacement policy must be able to generalize to unseen \emph{code paths} (i.e., sequences of accesses) from the same program, as there are exponentially many code paths and encountering them all during training is infeasible.
We test \ours's ability to generalize to new code paths by comparing it to the nearest neighbors baseline (\reffig{main_results}).
The performance of the nearest neighbors baseline shows that merely memorizing training code paths seen achieves near-optimal cache hit rates on simpler programs (e.g., gems, lbm), which just repeatedly execute the same code paths, but fails for more complex programs (e.g., mcf, Web Search), which exhibit highly varied code paths.
In contrast, \ours maintains high cache hit rates even on these more complex programs, showing that it can generalize to new code paths not seen during training.

Additionally, some of the programs require generalizing to new memory addresses and program counters at test time.
In mcf, 21.6\% of the test-time memory addresses did not appear in the training data,
and in Web Search, 5.3\% of the test-time memory addresses and 6\% of the test-time PCs did not appear in the training data (full details in \refapp{experiment_details}),
but \ours performs well despite this.

\subsection{Ablations}\label{sec:ablations}

\begin{figure}\center
    \includegraphics[width=\linewidth]{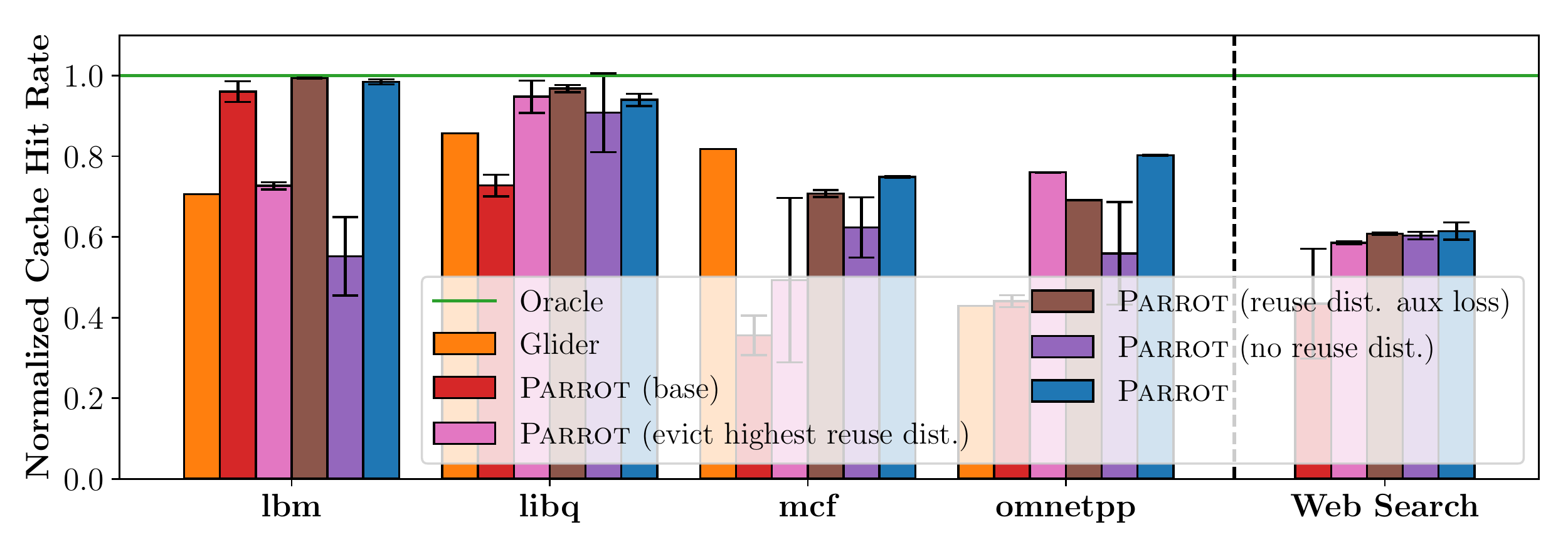}
    \caption{
        Comparison between different mechanisms of incorporating reuse distance into \ours.
Including reuse distance prediction in our full model (\ours) achieves 16.8\% higher normalized cache hit rates than ablating reuse distance prediction (\ours (no reuse dist.)).
}
    \label{fig:reuse_distance}
\end{figure}

Below, we ablate each of the following from \ours:
predicting reuse distance, on-policy training (DAgger), and ranking loss.
We evaluate on four of the most memory-intensive SPEC2006 applications (lbm, libq, mcf, and omnetpp) and Web Search and compare each ablation with Glider, Belady's, and two versions of \ours.
\ours is the full-sized model with no ablations.
\ours (base) is \ours's neural architecture, with all three additions ablated.
Comparing \ours(base) to Glider (e.g., \reffig{reuse_distance}) shows that in some programs (e.g., omnetpp and lbm), simply casting cache replacement as an IL problem with \ours's neural architecture is sufficient to obtain competitive performance, while in other programs, our additions are required to achieve state-of-the-art cache hit rates.

\paragraph{Predicting Reuse Distance.}

\reffig{reuse_distance} compares the following three configurations to show the effect of incorporating reuse distance information: (i)~\ours (no reuse dist.), where reuse distance prediction is ablated, (ii)~\ours (evict highest reuse dist.), where our fully ablated model (\ours (base)) predicts reuse distance and directly evicts the line with the highest predicted reuse distance, and (iii)~\ours (reuse dist. aux loss), where our fully ablated model learns to predict reuse distance as an auxiliary task.

Comparing \ours (no reuse dist.) to \ours shows that incorporating reuse distance greatly improves cache hit rates.
Between different ways to incorporate reuse distance into \ours,
using reuse distance prediction indirectly as an auxiliary loss function (\ours (reuse dist. aux loss)) leads to higher cache hit rates than using the reuse distance predictor directly to choose which cache line to evict (\ours (evict highest reuse dist.)).
We hypothesize that in some cache states, accurately predicting the reuse distance for each line may be challenging, but ranking the lines may be relatively easy.
Since our reuse distance predictor predicts log reuse distances,
small errors may drastically affect which line is evicted when the reuse distance predictor is used directly.

\paragraph{Training with DAgger.}

\begin{figure}\center
    \includegraphics[width=\linewidth]{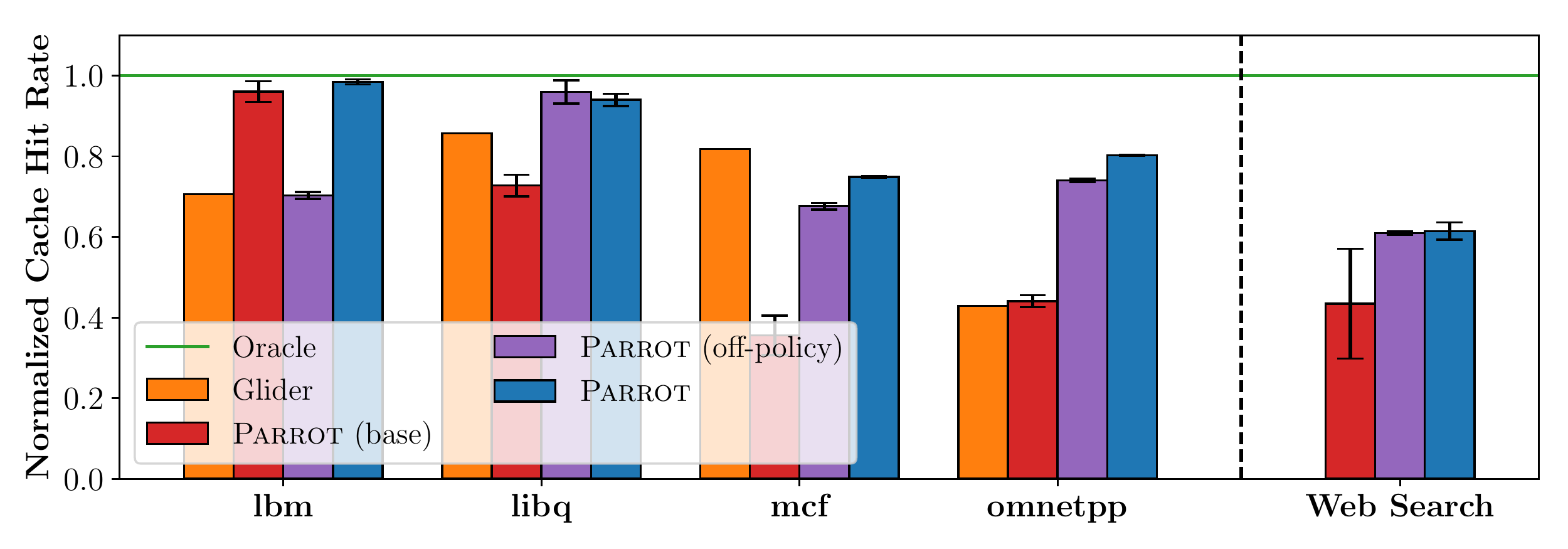}
    \caption{
        Ablation study for training with DAgger.
        Training with DAgger achieves 9.8\% higher normalized cache hit rates than training off-policy on the states visited by the oracle policy.
    }
    \label{fig:on_policy}
\end{figure}

\reffig{on_policy} summarizes the results when ablating training on-policy with DAgger.
In theory, training off-policy on roll-outs of Belady's should lead to compounding errors,
as the states visited during training under Belady's differ from those visited during test time. Empirically, we observe that this is highly program-dependent.
In some programs, like mcf or Web Search, training off-policy performs as well or better than training on-policy,
but in other programs, training on-policy is crucial.
Overall, training on-policy leads to an average 9.8\% normalized cache hit rate improvement over off-policy training.

\paragraph{Ranking Loss.}

\begin{figure}\center
    \includegraphics[width=\linewidth]{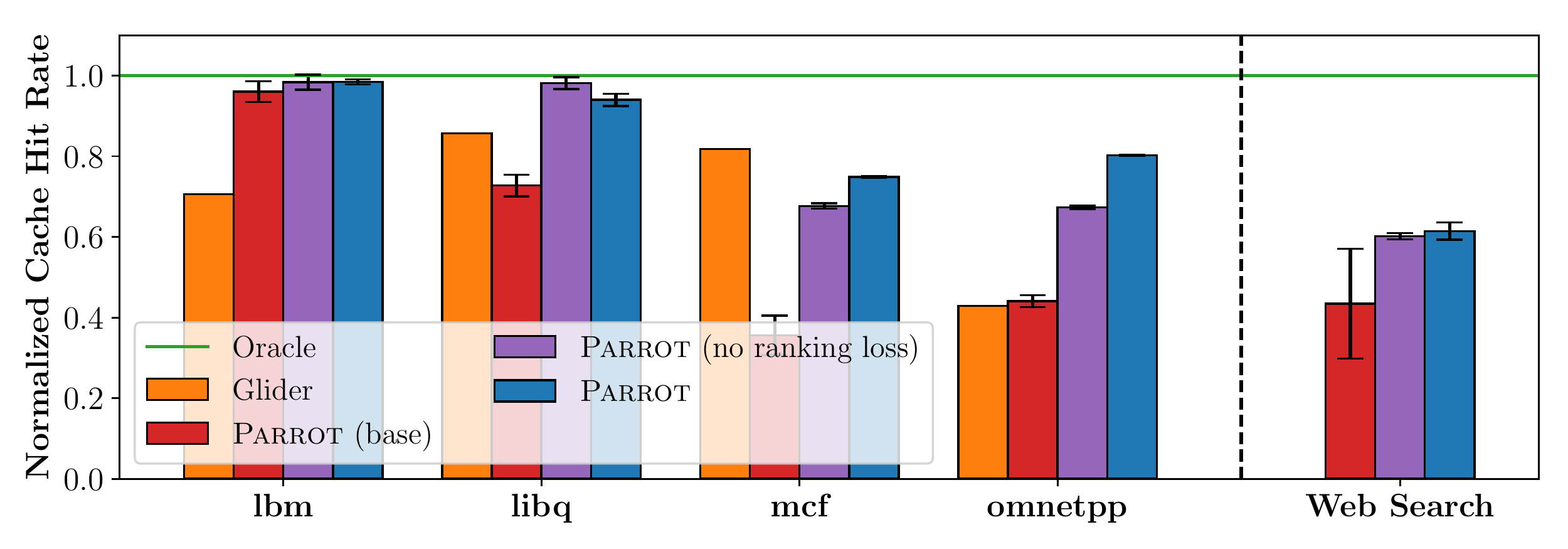}
    \caption{
        Ablation study for our ranking loss.
        Using our ranking loss improves normalized cache hit rate by 3.5\% over a LL loss.
    }
    \label{fig:ranking_loss}
\end{figure}

\reffig{ranking_loss} summarizes the results when ablating our ranking loss.
Using our ranking loss over a log-likelihood (LL) loss introduces some bias,
as the true optimal policy places all its probability on the line with the highest reuse distance.
However, our ranking loss better optimizes cache hit rates,
as it more heavily penalizes evicting lines with lower reuse distances, which lead to misses.
In addition, a distillation perspective of our loss, where the teacher network is an exponentially-smoothed version of Belady's with the probability of evicting a line set as proportional to $\exp(\textrm{reuse distance})$,
suggests that our ranking loss provides greater supervision than LL.
Tuning a temperature on the exponential smoothing of Belady's could interpolate between less bias and greater supervision.
Empirically, we observe that our ranking loss leads to an average 3.5\% normalized cache hit rate improvement over LL.

\subsection{History Length}
\label{sec:history}

One key question is: how much past information is needed to accurately approximate Belady's?
We study this by varying the number of past accesses that \ours attends over ($H$) from 20 to 140.
In theory, \ours's LSTM hidden state could contain information about \emph{all} past accesses,
but the LSTM's memory is limited in practice.

\begin{figure}\center
    \includegraphics[width=\linewidth]{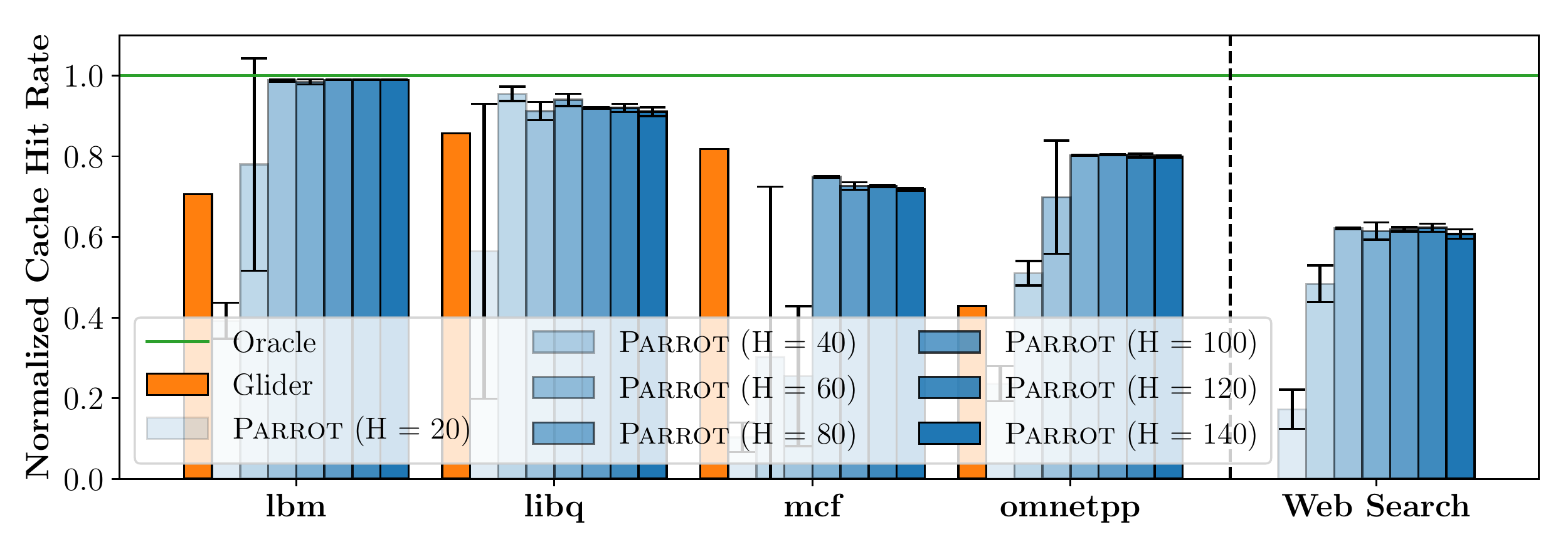}
    \caption{
        Performance of \ours trained with different numbers of past accesses ($H$).
        As the number of past accesses increases, normalized cache hit rates improve, until reaching a history length of 80.
        At that point, additional past accesses have little impact.
    }
    \label{fig:history_length}
\end{figure}

The results are summarized in \reffig{history_length}.
We observe that the past accesses become an increasingly better predictor of the future as the number past accesses increase, until about 80.
After that point, more past information doesn't appear to help approximate Belady's.
Interestingly, \citet{shi2019applying} show that Glider experiences a similar saturation in improvement from additional past accesses, but at around 30 past accesses.
This suggests that learning a replacement policy end-to-end with \ours can effectively leverage more past information than simply predicting whether a cache line is cache-friendly or cache-averse.

 \section{Related Work}\label{sec:related_work}

\paragraph{Cache Replacement.}

Traditional approaches to cache replacement rely on heuristics built upon intuition for cache access behavior.
LRU is based on the assumption that most recently used lines are more likely to be reused.
More sophisticated policies target a handful of manually classified access patterns based on simple counters~\cite{qureshi2007adaptive,jaleel2010high} or try to predict instructions that tend to load zero-reuse lines based on a table of saturating counters~\cite{wu2011ship,khan2010sampling}.

Several recent approaches instead focus on \emph{learning} cache replacement policies.
\citet{wang2019learning} also cast cache replacement as learning over a Markov decision process, but apply reinforcement learning instead of imitation learning, which results in lower performance.
More closely related to ours are Hawkeye~\cite{jain2016back} and Glider~\cite{shi2019applying},
which also learn from Belady's.
They train a binary classification model based on Belady's to predict if a line is cache-friendly or cache-averse, but rely on a traditional replacement heuristic to determine which line to evict when several lines are cache-averse.
Relying on the traditional heuristic to produce the final eviction decisions heavily constrains the expressivity of the policy class they learn over, so that even the best policy within their class of learnable policies may not accurately approximate Belady's, yielding high cache miss rates for some access patterns.

In contrast, our work is the first to propose learning a cache replacement policy end-to-end with imitation learning. Framing cache replacement in this principled framework is important as much prior research has resorted to heuristics for hill climbing specific benchmarks. In addition, learning end-to-end enables us to optimize over a highly expressive policy class, achieving high cache hit rates even on complex and diverse access patterns.

\paragraph{Imitation Learning.}

Our work builds on imitation learning (IL) techniques~\citep{ross2014reinforcement, sun2017deeply},
where the goal is to approximate an expert policy.
Our setting exhibits two distinctive properties:
First, in our setting, the expert policy (Belady's) can be queried at any state during training.
the oracle policy (Belady's) can be cheaply queried at any state during training, which differs from a body of IL work~\citep{vecerik2017leveraging, hosu2016playing, hester2018deep} focusing on learning with limited samples from an expensive expert (e.g., a human).
The ability to arbitrarily query the oracle enables us to avoid compounding errors with DAgger~\citep{ross2011reduction}.
Second, the distribution over actions of the oracle policy is available, enabling more sophisticated loss functions.
Prior work~\citep{sabour2018optimal, choudhury2017adaptive} also studies settings with these two properties, although in different domains.
\citet{sabour2018optimal} shows that an approximate oracle can be computed in some natural-language sequence generation tasks; \citet{choudhury2017adaptive} learns to imitate an oracle computed from data only available during training, similar to Belady's, which requires future information. \section{Conclusion and Future Directions}
\label{sec:discussion}

We develop a foundation for learning end-to-end cache replacement policies with imitation learning,
which significantly bridges the gap between prior work and Belady's optimal replacement policy.
Although we evaluate our approach on CPU caches, due to the popularity of SPEC2006 as a caching benchmark,
we emphasize that our approach applies to other caches as well, such as software caches, databases, and operating systems.
Software caches may be an especially promising area for applying our approach, as they tolerate higher latency in the replacement policy and implementing more complex replacement policies is easier in software.
We highlight two promising future directions:

First, this work focuses on the ML challenges of training a replacement to approximate Belady's and does not explore the practicality of deploying the learned policy in production,
where the two primary concerns are the memory and latency overheads of the policy.
To address these concerns, future work could investigate model-size reduction techniques, such as distillation~\citep{hinton2015distilling}, pruning~\citep{janowsky1989pruning, han2015deep, sze2017efficient}, and quantization, as well as domains tolerating greater latency and memory use, such as software caches.
Additionally, cache replacement decisions can be made at any time between misses to the same set, which provides a reasonably long latency window (e.g., on the order of seconds for software caches) for our replacement policy to make a decision.
Furthermore, the overall goal of cache replacement is to minimize latency.
While minimizing cache misses minimizes latency to a first approximation, cache misses incur variable amounts of latency~\citep{qureshi2006case}, which could be addressed by fine-tuning learned policies to directly minimize latency via reinforcement learning.

Second, while Belady's algorithm provides an optimal replacement policy for a single-level cache, there is no known optimal policy for multiple levels of caches (as is common in CPUs and web services). This \textit{hierarchical cache replacement} policy is a ripe area for deep learning and RL research, as is exploring the connection between cache replacement and prefetching, as they both involve selecting the optimal set of lines to be present in the cache. Cache replacement is backward looking (based on the accesses so far) while prefetching is forward looking (predicting future accesses directly~\citep{hashemi2018learning, shi2020}).

To facilitate further research in this area, we release a Gym environment for cache replacement, which easily extends to the hierarchical cache replacement setting, where RL is required as the optimal policy is unknown.
We find cache replacement an attractive problem for the RL\slash IL communities, as it has significant real-world impact and data is highly available, in contrast to many current benchmarks that only have one of these two properties.
In addition, cache replacement features several interesting challenges:
rewards are highly delayed, as evicting a particular line may not lead to a cache hit\slash miss until thousands of timesteps later;
the semantics of the action space dynamically changes, as the replacement policy chooses between differing cache lines at different states;
the state space is large (e.g., 100,000s of unique addresses) and some programs require generalizing to new memory addresses at test time, not seen during training, similar to the rare words problem~\citep{luong2014addressing} in NLP;
and as our ablations show, different programs exhibit wildly different cache access patterns, which can require different techniques to address.
In general, we observe that computer systems exhibit many interesting machine learning (ML) problems, but have been relatively inaccessible to the ML community because they require sophisticated systems tools.
We take steps to avoid this by releasing our cache replacement environment.

\paragraph{Reproducibility.}
Code for \ours and our cache replacement Gym environment is available at
\url{https://github.com/google-research/google-research/tree/master/cache_replacement}. \section*{Acknowledgements}
We thank Zhan Shi for insightful discussions and for providing the results for Glider, which we compare against.
We also thank Chelsea Finn, Lisa Lee, and Amir Yazdanbakhsh for their comments on a draft of this paper.
Finally, we thank the anonymous ICML reviewers for their useful feedback, which helped improve this paper.

This material is based upon work supported by the National Science Foundation Graduate Research Fellowship under Grant No. DGE-1656518.

\bibliography{main}
\bibliographystyle{icml2020}

\newpage

\clearpage{}\appendix
\section{Architecture Details}\label{sec:architecture_details}

Our model is implemented in PyTorch \citep{NEURIPS2019_9015} and is optimized with the Adam optimizer \citep{kingma2014adam}.

\paragraph{Embeddings.}

In the full-sized model, to embed memory addresses and cache lines,
we train a unique embedding for each unique memory address observed during training,
sharing the same embedder across memory addresses and cache lines.
We similarly embed PCs.

Concretely,
we initialize an embedding matrix $W^m \in \mathbb{R}^{(n_m + 1) \times d_m}$ via Glorot uniform initialization \citep{glorot2010understanding}, 
where $n_m$ is set to the number of unique memory addresses in the training set and $d_m$ is the dimension of the embedding.
Then, each unique memory address $m$ is dynamically assigned a unique id 1 to $n_m$, and its embedding $e(m)$ is set to the $i$-th row of the embedding matrix $W^m$.
At test time, all memory addresses unseen during training are mapped to a special \emph{UNK} embedding, equal to the last row of $W^m$.
We embed PCs with a similar embedding matrix $W^p \in \mathbb{R}^{(n_p + 1) \times d_p}$.

\paragraph{Attention.}

After computing embeddings (with either the full-sized model or the byte embedder)
$e(l_1), \ldots, e(l_W)$ for each line in the cache state
and hidden states $[h_{t - H + 1}, \ldots, h_t]$ representing the past $H$ accesses,
we compute a context $g_w$ for each cache line by attending over the hidden states with the line embeddings as queries as follows:

\begin{enumerate}
    \item
        Following \citet{vaswani2017attention}, we compute positional embeddings $e(-H + 1), \ldots, e(0)$, where $e(\mathrm{pos}) \in \mathbb{R}^{d_\mathrm{pos}}$ and:
      
        \begin{align*}
            e(\mathrm{pos})_{2i} &= \sin\left(\frac{\mathrm{pos}}{10000^{2i / d_{\mathrm{pos}}}}\right) \\
            e(\mathrm{pos})_{2i + 1} &= \cos\left(\frac{\mathrm{pos}}{10000^{2i / d_{\mathrm{pos}}}}\right).
        \end{align*}
        
        We concatenate these positional embeddings with the hidden states to encode how far in the past each access is:
        $[(h_{t - H + 1}; e(-H + 1)), \ldots, (h_t; e(0))]$.
        Although in theory, the LSTM hidden states can encode positions, we found that explicitly concatenating a positional embedding helped optimization.
    
    \item 
        We apply General Attention \citep{luong2015effective} with each cache line embedding as the query and the concatenated hidden states and positional embeddings as keys:
        
        \begin{align*}
            \alpha_i &= \textrm{softmax}(e(l_w)^T W_e h_{t - H + i}) \\
            g_w &= \sum_{i = 1}^H \alpha_i h_{t - H + i}.
        \end{align*}
        
        The matrix $W_e \in \mathbb{R}^{d_m \times (d_\mathrm{LSTM} + d_\mathrm{pos})}$ is learned and $g_1, \ldots, g_W$ can be computed in parallel \citep{vaswani2017attention}.
        
\end{enumerate}

\section{Experimental Details}\label{sec:experiment_details}

\subsection{Detailed Results}

To provide further insight into our learned policy, we also report results on two additional metrics:

\begin{itemize}
    \item
        \textbf{Top-$K$ Accuracy}: The percentage of the time that the optimal line to evict according to Belady's is in the top-$K$ lines with highest probability of eviction under our learned policy.
        This indicates how frequently our learned policy is outputting decisions like those of Belady's.
        We report top-$1$ and top-$5$ accuracy.
        Note that since each cache set in the last-level cache can only hold $16$ lines, the top-$16$ accuracy is 100\%.
    \item
        \textbf{Reuse Distance Gap}: The average difference between the reuse distance of the optimal line to evict $l^*$ and the reuse distance of the line evicted by \ours, i.e., $d_t(l^*) - d_t(l_w)$, where $w = \arg\max_i \pi(i \mid s_t)$.
        This metric roughly captures how sub-optimal the decision made by \ours is at each timestep,
        as evicting a line with a smaller reuse distance is more likely to lead to a cache miss.
        A policy with a reuse distance gap of $0$ is optimal, while a high reuse distance gap leads to more cache misses.
\end{itemize}

We report results with of our full model with this metric in \reftab{additional_results}.
In mcf, libquantum, and lbm, our replacement policy frequently chooses to evict the same cache line as Belady's with a top-1 accuracy of over $75\%$ and a top-5 accuracy close to $100\%$.
In other programs (e.g., omnetpp, Web Search), our replacement policy's top-1 and top-5 accuracies are significantly lower, even though its normalized cache hit rate in these programs is similar to its normalized cache hit rate in mcf.
Intuitively, this can occur when several cache lines have similar reuse distances to the reuse distance of the optimal cache line, so evicting any of them is roughly equivalent.
Thus top-$K$ accuracy is an interesting, but imperfect metric.
Note that this is the same intuition behind our ranking loss, which roughly measures the relative suboptimality of a line as a function of the suboptimality of its reuse distance.

The differences in the reuse distance gaps between the programs emphasize the differences in the program behaviors and roughly indicate how frequently cache lines are reused in each program.
For example, \ours achieves wildly different average reuse distance gaps, while maintaining similar normalized and raw cache hit rates (\reftab{raw_cache_hit_rates}) in omnetpp and mcf,
due to differences in their access patterns.
In omnetpp, evicting a line with a reuse distance 100's of accesses smaller than the reuse distance of the optimal line sometimes does not lead to a cache miss,
as both lines may eventually be evicted before being reused anyway.
On the other hand, in mcf, evicting a line that is used only slightly earlier than the optimal line more frequently leads to cache misses.

\begin{table}
    \centering\small
    \caption{
        Mean top-1/5 accuracy and reuse distance gap of \ours, averaged over 3 seeds with single standard deviation intervals.
    }
    \vspace{5pt}
    \resizebox{\columnwidth}{!}{\renewcommand{\arraystretch}{1.2}
        \begin{tabular}{ lccc }
          \toprule
          Program & Top-1 Acc. (\%) & Top-5 Acc. (\%) & Reuse Dist. Gap \\
          \midrule
          astar      & $17.2 \pm 1.1$ & $50.8 \pm 1.8$ & $20065.8 \pm 6433.1$ \\
          bwaves     & $20.9 \pm 3.2$ & $49.1 \pm 3.0$ & $1356.9 \pm 653.5$ \\
          bzip       & $21.2 \pm 0.9$ & $44.3 \pm 2.5$ & $478.1 \pm 39.5$\\
          cactusadm  & $91.3 \pm 1.7$ & $98.3 \pm 0.2$ & $2.0 \pm 0.1$ \\
          gems       & $15.0 \pm 0.6$ & $45.5 \pm 2.4$ & $46.4 \pm 3.8$ \\
          lbm        & $83.6 \pm 2.4$ & $98.0 \pm 0.4$ & $2.1 \pm 0.3$ \\
          leslie3d   & $88.5 \pm 0.2$ & $98.1 \pm 0.2$ & $1.9 \pm 0.0$ \\
          libquantum & $83.2 \pm 3.9$ & $97.3 \pm 0.2$ & $2.2 \pm 0.6$ \\
          mcf        & $75.2 \pm 0.5$ & $87.6 \pm 0.3$ & $6.4 \pm 0.4$ \\
          milc       & $45.2 \pm 2.2$ & $65.6 \pm 2.3$ & $39.8 \pm 5.1$ \\
          omnetpp    & $35.0 \pm 0.4$ & $65.1 \pm 0.4$ & $533.4 \pm 8.0$ \\
          sphinx3    & $67.0 \pm 2.0$ & $88.4 \pm 1.9$ & $39.0 \pm 8.5$ \\
          xalanc     & $40.4 \pm 2.2$ & $93.4 \pm 0.7$ & $34805.1 \pm 14760.2$ \\
          \hline
          Web Search & $31.8 \pm 0.8$ & $77.5 \pm 3.6$ & $4012.5 \pm 413.9$ \\
          \bottomrule
        \end{tabular}
    }
  \label{tab:additional_results}
\end{table}

\subsection{Hyperparameters}

The following shows the values of all the hyperparameters we used in all our final experiments.
We ran our final experiments with the bolded values and tuned over the non-bolded values.
These values were used in all of our final experiments, including the ablations,
except the history length experiments (\refsec{history}), where we varied the history length $H$.
\begin{itemize}
    \item Learning rate: (\textbf{0.001}, 0.003)
    \item Address embedding dimension ($d_m$): \textbf{64}
    \item PC embedding dimension ($d_p$): \textbf{64}
    \item PC embedding vocab size ($n_p$): \textbf{5000}
    \item Position embedding dimension ($d_\mathrm{pos}$): \textbf{128}
    \item LSTM hidden size ($d_\mathrm{LSTM}$): \textbf{128}
    \item Frequency of recollecting $B$: (\textbf{5000}, 10000)
    \item History Length ($H$): (20, 40, 60, \textbf{80}, 100, 120, 140)
\end{itemize}

We used the same hyperparameter values in all 5 programs (omnetpp, mcf, libq, lbm, and Web Search) we evaluated on,
where the address embedding vocab size $n_m$ was set to the number of unique addresses seen in the train set of each program
(see \reftab{program-details}).
For most hyperparameters, we selected a reasonable value and never changed it.
We tuned the rest of the hyperparameters exclusively on the validation set of omnetpp.

\subsection{Program Details}

\begin{table*}
    \centering\small
    \caption{
         Program details in terms of the number of last-level cache accesses and unique addresses/PCs.
         `Train' and `Test' show the number of unique addresses and PCs appearing in each domain at train and test time,
         whereas `Unseen Test' indicates the number of addresses and PCs appearing at test time, but not train time.
         The given percentages indicate what portion of all test accesses had unseen addresses or PCs.
    }
    \vspace{5pt}
    \renewcommand{\arraystretch}{1.1}
    \begin{tabular}{ lccccccc }
      \toprule
      & & \multicolumn{2}{c}{Train} & \multicolumn{2}{c}{Test} & \multicolumn{2}{c}{Unseen Test} \\
      \cmidrule(lr){3-4}\cmidrule(lr){5-6}\cmidrule(l){7-8}
      Program & Cache Accesses & Addresses & PCs & Addresses & PCs & Addresses & PCs \\
      \midrule
      astar & 879,040 & 13,047 & 25 & 8,235 & 9 & 46 (0.6\%) & 0 (0\%) \\
      bwaves & 2,785,280 & 443,662 & 436 & 209,231 & 155 & 12 (0.1\%) & 38 (5.8\%) \\
      bzip & 3,899,840 & 6,086 & 415 & 3,571 & 28 & 0 (0\%) & 0 (0\%) \\
      cactusadm & 1,759,040 & 298,213 & 243 & 83,046 & 191 & 0 (0\%) & 0 (0\%) \\
      gems & 7,298,880 & 423,706 & 5,305 & 363,313 & 1,395 & 1 (0\%) & 0 (0\%) \\
      lbm & 10,224,960 & 206,271 & 44 & 206,265 & 34 & 0 (0\%) & 0 (0\%) \\
      leslie3d & 6,956,160 & 38,507 & 1,705 & 38,487 & 1,663 & 0 (0\%) & 0 (0\%) \\
      libquantum & 4,507,200 & 16,394 & 14 & 16,386 & 5 & 0 (0\%) & 0 (0\%) \\
      mcf & 4,143,360 & 622,142 & 73 & 81,666 & 67 & 77,608 (21.6\%) & 0 (0\%) \\
      milc & 3,048,000 & 203,504 & 254 & 90,933 & 82 & 0 (0\%) & 2 (0.5\%) \\
      omnetpp & 3,414,720 & 16,912 & 402 & 14,079 & 315 & 275 (0.6\%) & 3 (1.0\%) \\
      sphinx3 & 2,372,800 & 20,629 & 528 & 4,586 & 199 & 0 (0\%) & 1 (0\%) \\
      xalanc & 4,714,240 & 15,515 & 217 & 11,600 & 161 & 507 (1\%) & 5 (0\%) \\
      \hline
      Web Search & 3,636,800 & 241,164 & 32,468 & 66,645 & 15,893 & 10,334 (5.3\%) & 948 (6\%) \\
      \bottomrule
    \end{tabular}
  \label{tab:program-details}
\end{table*}

\reftab{program-details} reports the number of unique addresses and PCs contained in the train\slash test splits of each program,
including the number of unique addresses and PCs in the test split that were not in the train split.
Notably, in some programs, new addresses and PCs appear at test time, that are not seen during training, requiring the replacement policy to generalize.

In \reftab{program-details}, we also report the total number of last-level cache accesses collected for each of the five programs in our 50s collection interval.
These accesses were split into the 80\% train, 10\% validation, and 10\% test sets.
Since different programs exhibited varying levels of cacheability at the L1 and L2 cache levels, different numbers of last-level cache accesses resulted for each program.
These varying numbers also indicate how different programs exhibit highly different behavior.

\subsection{Randomly Chosen Cache Sets}

We randomly chose 64 sets and collected accesses to those sets on the last-level cache.
The 64 randomly chosen sets were: 6, 35, 38, 53, 67, 70, 113, 143, 157, 196, 287, 324, 332, 348, 362, 398, 406, 456, 458, 488, 497, 499, 558, 611, 718, 725, 754, 775, 793, 822, 862, 895, 928, 1062, 1086, 1101, 1102, 1137, 1144, 1175, 1210, 1211, 1223, 1237, 1268, 1308, 1342, 1348, 1353, 1424, 1437, 1456, 1574, 1599, 1604, 1662, 1683, 1782, 1789, 1812, 1905, 1940, 1967, and 1973.\clearpage{}

\end{document}